\documentclass[letterpaper, 10 pt, journal, twoside]{ieeetran}  % Comment this line out if you need a4paper

\IEEEoverridecommandlockouts                              % This command is only needed if 
\usepackage{graphics} % for pdf, bitmapped graphics files
\usepackage{subfiles}
\usepackage{csquotes}
\usepackage{gensymb}
\usepackage{pifont}
\usepackage{textcomp}
\usepackage[table]{xcolor}
\usepackage{todonotes}
\usepackage{pdfpages}
\usepackage{blindtext}
\usepackage{soul}
\usepackage{hyperref}
\usepackage{graphicx}
\usepackage{mathtools}
\usepackage{multirow}
\usepackage{booktabs}
\graphicspath{{images/}{../images/}}
\usepackage{subcaption}
\usepackage{caption}
\renewcommand{\baselinestretch}{0.975}
\usepackage[export]{adjustbox}
\usepackage{xcolor}

\usepackage{amsmath} 
\usepackage{amssymb} 
\usepackage{amsfonts}
\usepackage{bm} 
\usepackage{siunitx}
\usepackage{cleveref}
\usepackage[inline]{enumitem}

\usepackage{microtype}
\newcommand{\secref}[1]{Sec.~\ref{#1}}
\renewcommand{\eqref}[1]{Eq.~(\ref{#1})}
\newcommand{\figref}[1]{Fig.~\ref{#1}}
\newcommand{\tabref}[1]{Tab.~\ref{#1}}

\newcolumntype{P}[1]{>{\centering\arraybackslash}p{#1}}

\newcommand{\net}{AdaptLPS}

\definecolor{road}{RGB}{255,0,255}
\definecolor{building}{RGB}{255,200,0}
\definecolor{other-vehicle}{RGB}{0,0,255}
\definecolor{sidewalk}{RGB}{75,0,75}
\definecolor{vegetation}{RGB}{0,175,0}
\newcommand{\colsq}[1]{(\textcolor{#1}{\rule{0.2cm}{0.2cm}})}
\newcommand{\rebuttal}[1]{\textcolor{black}{#1}}
\newcommand{\acceptance}[1]{\textcolor{black}{#1}}

\usepackage{tikz}
\usepackage{textcomp}
\usepackage{lipsum}

\title{
Unsupervised Domain Adaptation for LiDAR Panoptic Segmentation
}

\author{Borna Be\v{s}i\'{c}\textsuperscript{$\ast$}, Nikhil Gosala\textsuperscript{$\ast$}, Daniele Cattaneo, and Abhinav Valada% <-this % stops a space
\thanks{\textsuperscript{$\ast$}Authors contributed equally.}
\thanks{All authors are with the Department of Computer Science, University of Freiburg, Germany.}
}

\begin{document}

% Paper headers
\markboth{\copyright\ 2022 IEEE}
{\copyright\ 2022 IEEE} 
% Use only for final RAL version

\maketitle
% \pagestyle{empty}
% \thispagestyle{empty}
% Comment or remove these lines for final RAL version.

%%%%%%%%%%%%%%%%%%%%%%%%%%%%%%%%%%%%%%%%%%%%%%%%%%%%%%%%%%%%%%%%%%%%%%%%%%%%%%%%
\begin{abstract}
Scene understanding is a pivotal task for autonomous vehicles to safely navigate in the environment. Recent advances in deep learning enable accurate semantic reconstruction of the surroundings from LiDAR data. However, these models encounter a large domain gap while deploying them on vehicles equipped with different LiDAR setups which drastically decreases their performance.
Fine-tuning the model for every new setup is infeasible due to the expensive and cumbersome process of recording and manually labeling new data. Unsupervised Domain Adaptation (UDA) techniques are thus essential to fill this domain gap and retain the performance of models on new sensor setups without the need for additional data labeling. In this paper, we propose \net, a novel UDA approach for LiDAR panoptic segmentation that leverages task-specific knowledge and accounts for variation in the number of scan lines, mounting position, intensity distribution, and environmental conditions. We tackle the UDA task by employing two complementary domain adaptation strategies, data-based and model-based. While data-based adaptations reduce the domain gap by processing the raw LiDAR scans to resemble the scans in the target domain, model-based techniques guide the network in extracting features that are representative for both domains.
Extensive evaluations on three pairs of real-world autonomous driving datasets demonstrate that \net\ outperforms existing UDA approaches by up to 6.41~pp in terms of the PQ score.
\end{abstract}


%%%%%%%%%%%%%%%%%%%%%%%%%%%%%%%%%%%%%%%%%%%%%%%%%%%%%%%%%%%%%%%%%%%%%%%%%%%%%%%%
\section{Introduction}

\IEEEPARstart{A}{utonomous} vehicles (AVs) rely on accurate semantic understanding of their surroundings for reliable and safe operation. Scene segmentation is extensively used in various applications such as dynamic object removal~\cite{bevsic2020dynamic} and localization~\cite{boniardi2019robot} as it enables distinguishing points that belong to different objects and classes. It can be classified into three tasks, namely, semantic segmentation which predicts a class label for each point, instance segmentation which assigns a unique ID to points belonging to each object, and panoptic segmentation which combines both semantic and instance segmentation to yield a holistic output containing both \textit{stuff} and \textit{thing} classes. Panoptic segmentation has gained significant interest in the recent years due to its comprehensive description of the scene.

\begin{figure}
    \centering
     \includegraphics[width=0.8\linewidth]{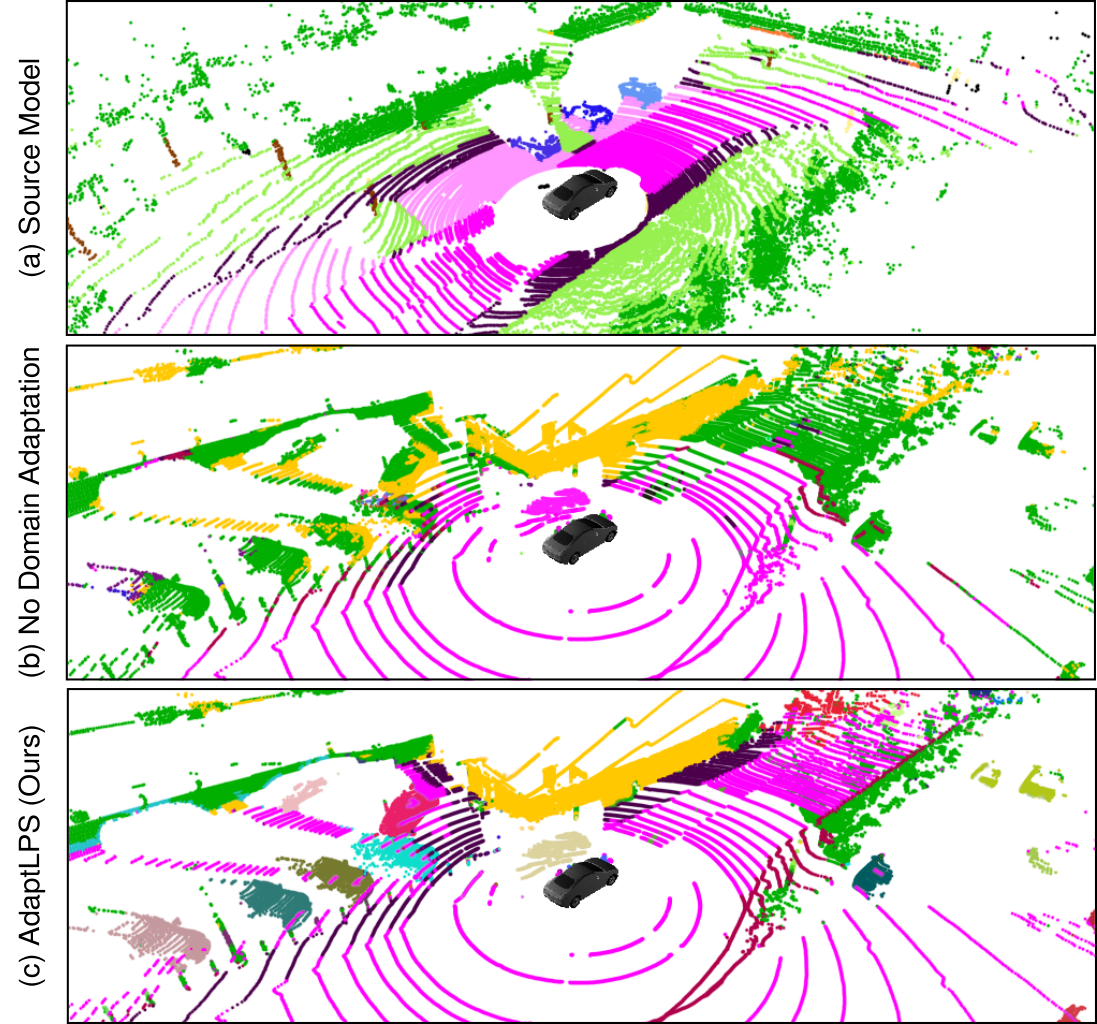}
     \caption{(a) State-of-the-art LiDAR panoptic segmentation networks achieve remarkable results when evaluated on scans from the same LiDAR setup used to train the model. (b) However, their performance drops when evaluated on scans from a different LiDAR setup. (c) Our UDA method minimizes this domain gap without any additional supervision and generates accurate panoptic segmentation outputs.}
     \label{fig:paper-teaser}
     \vspace{-3mm}
\end{figure}

State-of-the-art LiDAR panoptic segmentation approaches \cite{sirohi2021efficientlps,milioto2020lidar} use deep learning techniques that learn the characteristics of the training dataset and generate accurate panoptic predictions for samples drawn from the same distribution. However, evaluating the trained model on samples from a different domain, i.e., samples from a LiDAR having different mounting positions, number of scan lines, or intensity calibration, results in a significant performance drop due to the domain gap introduced by the notably different data distribution~(\figref{fig:paper-teaser}). This domain gap is more prominent in LiDAR point clouds due to the different position and depth estimates obtained for a given 3D point when a LiDAR is mounted at different positions and orientations on the AV. This domain gap is further widened due to the substantially different intensity distributions between LiDARs from different manufacturers. Since manual data annotation is extremely expensive and cumbersome, we aim to minimize the domain gap using Unsupervised Domain Adaptation (UDA) which adapts a model trained on a labeled dataset (source domain) to a new unlabeled dataset (target domain) without additional supervision.

Current state-of-the-art UDA approaches for LiDAR semantic segmentation~\cite{wu2018squeezesegv2,langer2020iros} minimize the domain gap by aligning second-order feature statistics of the source and target domains. However, statistics computed on the complete LiDAR scans are biased towards the most prevalent classes such as road and building, resulting in the inferior adaptation of the under-represented classes.
Moreover, existing approaches are not specifically designed for the task of panoptic segmentation, and therefore do not make use of essential task-specific information such as the knowledge of instances, which results in sub-optimal domain adaptation performance. 

In this work, we overcome the aforementioned limitations by proposing the first holistic UDA approach for LiDAR panoptic segmentation that comprises two independent domain adaptation strategies. 
Specifically, we split the task into \textit{data-based} and \textit{model-based} adaptation to independently reduce the domain gap in the raw LiDAR data and learned models respectively. The proposed data-based domain adaptation accounts for the disparity in the raw LiDAR data by
(i) transforming the poses of all LiDAR scans to a common reference frame,
(ii) simulating the characteristics of the target domain by generating semi-synthetic scans of the source data, and
(iii) mapping the intensities between the domains using an Intensity Mapping Network.
Our model-based domain adaptation, on the contrary, minimizes the domain gap in the learned models using the solution of an unbalanced optimal transport problem defined between multiple layers of the source and target feature maps. We augment this formulation with an instance-aware sampling paradigm to align features that are well distributed across classes and instances, and not dominated by few classes. Further, we also propose an efficient domain calibration strategy, PDC-Lite, to adapt the batch normalization parameters of the first layer to the target domain prior to the inference phase.
We perform extensive evaluations of our approach on three dataset pairs created using combinations of SemanticKITTI~\cite{behley2019iccv}, nuScenes~\cite{nuscenes2019}, and PandaSet~\cite{pandaset}, and demonstrate that our domain adaptation strategy outperforms the state-of-the-art by up to $6.41$ percentage points in terms of the PQ metric. Code and video of our proposed approach is publicly available at \url{http://lidar-panoptic-uda.cs.uni-freiburg.de}.
\section{Related Work}
\label{sec:related-work}

\noindent\textit{3D Unsupervised Domain Adaptation}: Existing machine learning approaches rely on the assumption that the training and testing samples are drawn from identically distributed datasets. However, when this assumption is violated, the trained model fails to generalize over the test dataset which results in a significant performance drop, and the discrepancy between the two distributions is referred to as the \textit{domain gap}. Often, the dataset on which we intend to evaluate our trained model, \textit{i.e.}, the target domain, is unlabeled which gives rise to UDA wherein a model trained on a labeled source dataset is adapted to a differently-distributed target dataset without additional supervision.
Multiple approaches address the challenge of UDA using geodesic correlation alignment (GCA)~\cite{morerio2018minimalentropy}, a minimal-entropy extension of Deep CORAL~\cite{dcoral}, to align the second-order feature statistics of the source and target domains~\cite{wu2018squeezesegv2, langer2020iros}. In addition to GCA, \cite{wu2018squeezesegv2} also proposes the use of progressive domain calibration~(PDC) to re-calibrate batch normalization layers on the target domain before evaluation, while~\cite{langer2020iros} generates semi-synthetic point clouds of the source data to better represent the characteristics of the raw data in the target domain. Further, \cite{alonso2020domain} minimizes the KL divergence between the source and target domains while simultaneously minimizing the entropy on the target domain. In contrast, Complete~\&~Label~\cite{Yi2021CVPR} uses a learning-based model to recover the underlying structures of 3D surfaces which serves as the canonical domain across different domains, while LidarNet~\cite{jiang2021lidarnet} uses a boundary-aware CycleGAN~\cite{CycleGAN2017} to minimize the domain gap between unpaired samples.

{\parskip=5pt
\noindent\textit{3D Panoptic Segmentation}: Existing approaches can be categorized into two groups, namely, \textit{proposal-based} and \textit{proposal-free} methods. Proposal-based approaches fuse the outputs from separate semantic and instance prediction stages to generate the panoptic output. The current state-of-the-art 3D panoptic segmentation approach, EfficientLPS~\cite{sirohi2021efficientlps}, follows this paradigm and employs scale-invariant semantic and instance distance-dependent heads along with a panoptic periphery loss.
PanopticTrackNet~\cite{hurtado2020mopt} further enforces temporal consistency of object instance IDs by performing object tracking along with panoptic segmentation. In contrast, proposal-free approaches predict point-wise semantic labels before clustering them into instances using a voting scheme or by employing a pixel-pair affinity pyramid~\cite{gao2019ssap}. Milioto~\textit{et~al.}~\cite{milioto2020lidar} project the points onto a 2D image and predict their offsets to object centroids before grouping them in a post-processing step. Panoster~\cite{gasperini2020panoster} employs KPConv together with a learnable clustering algorithm that removes the need for an additional post-processing stage to group points into instances. SMAC-Seg~\cite{li2021smac} uses a learnable multi-directional clustering along with a centroid-aware loss function to differentiate between object clusters.}

In this work, we address three limitations of existing domain adaptation approaches, namely, (i) the lack of an outlier handling mechanism in statistic alignment-based approaches which corrupts both the first and second-order statistics when the source and target mini-batches are not sufficiently similar, (ii) the inability of the existing approaches to account for local feature statistics such as those of instances and small objects, and (iii) long training times and extremely volatile convergence conditions of adversarial learning-based approaches which makes training such models very challenging. Accordingly, we formulate our domain adaptation as an unbalanced optimal transport problem that is robust against outliers and allows for a weighted mapping between similar features across the source and target mini-batches which helps preserve the local structure of the feature space. Moreover, being non-adversarial, it is free from the problems faced by such approaches resulting in a relatively compact training protocol.

\section{Technical Approach}
\label{sec:techincal-approach}

\begin{figure*}
    \centering
     \includegraphics[width=0.85\linewidth,trim={0 1.4cm 0 0},clip]{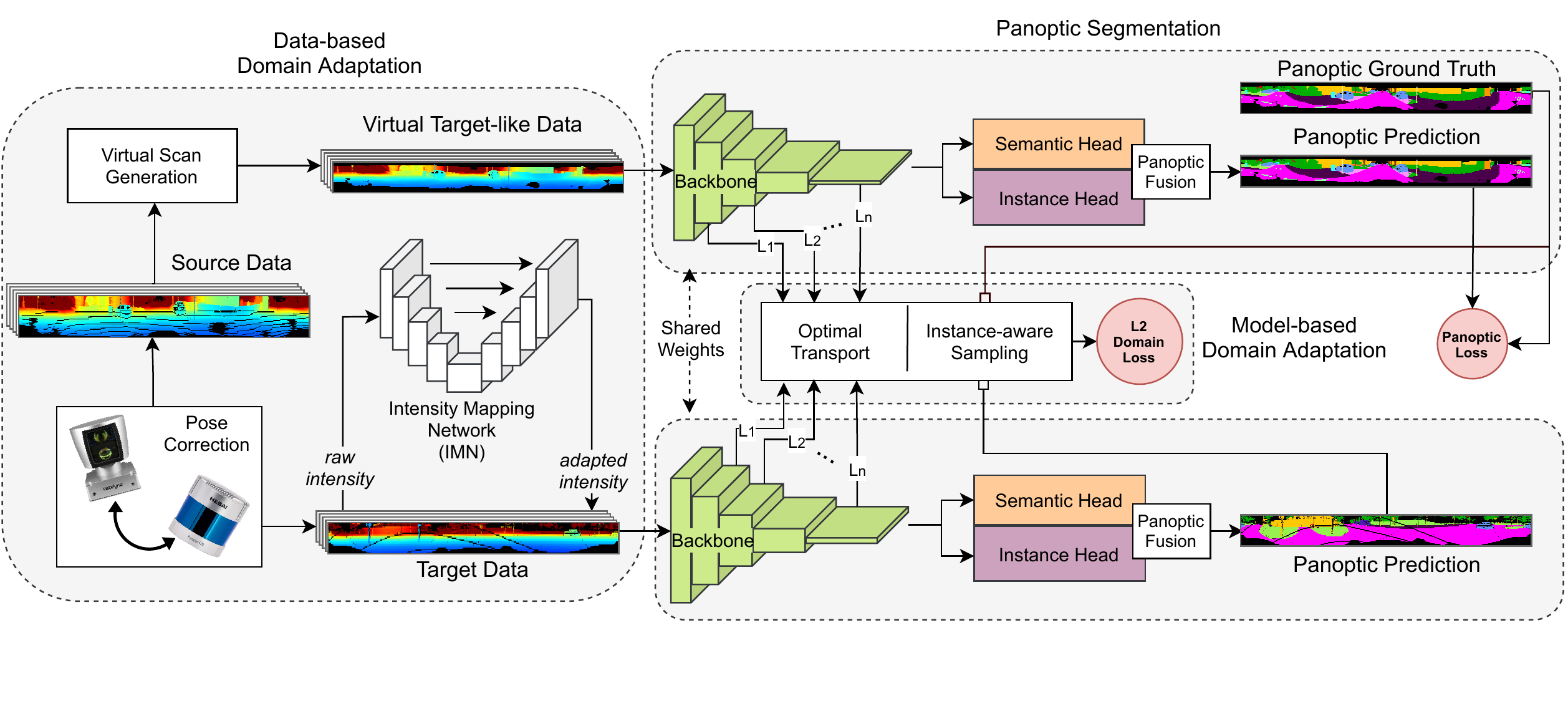}
     \caption{Illustration of our \net\ domain adaptation approach. We employ data-based (pose correction, virtual scan generation, and intensity mapping) and model-based (optimal transport and instance-aware sampling) domain adaptation strategies to reduce the domain gap between source and target datasets.}
     \label{fig:architecture}
     \vspace{-3mm}
\end{figure*}

In this section, we describe our proposed \net\ for UDA of LiDAR panoptic segmentation. Our approach, illustrated in \figref{fig:architecture}, comprises two domain adaptation strategies, namely, \textit{Data-based} and \textit{Model-based}. The data-based domain adaptation minimizes the domain gap between the raw LiDAR scans from the source and target domains using three techniques, namely, 
(i) \textit{pose correction} which accounts for the different mounting positions of LiDARs, 
(ii) \textit{virtual scan generation} which simulates the point cloud from the target domain using the source domain, and
(iii) \textit{intensity mapping} which learns a residual to map the LiDAR intensities between the source and target domains. 
In contrast, model-based domain adaptation minimizes the domain gap between the panoptic segmentation models by aligning the source and target distributions using our novel \textit{multi-scale feature-space optimal transport} augmented with \textit{instance-aware sampling}. During inference, we further reduce the domain gap using our efficient \textit{PDC-Lite} technique which re-calibrates the first batch normalization layer to remove any bias and variance accumulated from the source domain.
\vspace{-1mm}

\subsection{Data-Based Domain Adaptation}
\subsubsection{Pose Correction}
\label{sec:pose_correction}

The 3D position of points generated by a LiDAR sensor depends on its mounting position on the vehicle.
Although points are always expressed with respect to the LiDAR reference frame, their offset and rotation rely on the orientation as well as the height at which the sensor is placed.
Since the difference in these parameters can be quite significant between two distinct setups, we correct the poses of each LiDAR scan using our pose correction module. Specifically, we first detect the road plane by regressing its coefficients using RANSAC~\cite{fischler1981random}. To ensure that the detected plane is plausibly the road plane, we only use points below the sensor's height and constrain the plane normal to be as close as possible to the Z-axis. We rely on the assumption that the sensor's base should be approximately parallel to the ground or, alternatively, that an extrinsic calibration between the LiDAR and the vehicle is available. Once the road plane is estimated, we rotate the LiDAR to be parallel to it and set its height w.r.t. the road plane to a fixed value of \SI{1.75}{\meter}.

\subsubsection{Virtual Scan Generation}

Most state-of-the-art panoptic segmentation networks operate on range images $\bm{RI} \in \mathbb{R}^{C \times H \times W}$ of height $H$ and width $W$ with $ C = 5 $ channels, namely, range, $ x $, $ y $, $ z $ and intensity.
Accordingly, we project the 3D LiDAR points onto a 2D image plane having a fixed width of \SI{1024}{px} and height equaling the number of LiDAR scan lines.
The varying number of scan lines between the LiDAR sensors introduces yet another domain gap between the source and target domains due to the different fields-of-view and resolutions observed by the LiDAR sensors.
We address this domain gap by first generating realistic-looking, semi-synthetic range images from the source data to resemble the target data, and then train the model on this semi-synthetic dataset instead of the raw source point clouds. 
\rebuttal{We generate the semi-synthetic images by following an approach similar to the one proposed by Langer~\textit{et~al.}~\cite{langer2020iros} with a few crucial differences that make our method more scalable and accurate.
We first build a map of the environment by aggregating scans along each trajectory using points belonging to either \textit{stuff} classes or static instances. We observe those points are dense enough and therefore do not require mesh generation.}
We then generate meshes from points belonging to dynamic instances using 2D Delaunay triangulation, and uniformly sample a large number of points from each of the generated meshes. \rebuttal{This helps us achieve topologically correct triangle meshes that do not contain any holes.} We generate these meshes to densify the otherwise extremely sparse representation of dynamic objects in each synthetic sample. Further, we estimate the intensity of the newly sampled points using $3$-nearest-neighbor regression, where the neighbor's intensities are weighted by the inverse of their distances from the virtual LiDAR point. 
\rebuttal{Finally, we employ ray casting using the known LiDAR sensor model characteristics to project the combined point cloud onto a 2D image having the same vertical dimension as the target data}.\looseness=-1 

\subsubsection{Intensity Mapping Network (IMN)}

Different LiDAR sensors often employ different intensity calibrations which further increases the domain gap between the source and target domains. We account for this disparity by introducing an Intensity Mapping Network (IMN) that operates on target data and aligns its intensity distribution with that of the source domain. IMN predicts a residual which when added to the intensity channel of the original target data, generates the transformed target intensity channel. Mathematically, $\bm{I}_{t \rightarrow s} = \bm{I}_{t} + \text{IMN}\left(\bm{I}_{t}\right),$
where $ I_{t} $ is the original target data intensity and $ I_{t \rightarrow s} $ is the target intensity mapped to the source domain. To this end, we model IMN using a simple encoder-decoder architecture. We train this model in an end-to-end manner together with the main panoptic segmentation network.
\vspace{-2mm}

\subsection{Model-based Domain Adaptation}

\subsubsection{Multi-Scale Feature-Space Optimal Transport (MS-FSOT)}

Optimal transport establishes a framework for comparing two probability measures, $\mu$ over space~$\mathcal{X}$ and $\nu$ over space~$\mathcal{Y}$, and provides a solution for moving the mass from one distribution to the other. The transportation of mass is done with respect to a cost function $c$ which represents the difficulty in moving the mass from $ x \in \mathcal{X} $ to $ y \in \mathcal{Y} $. In short, optimal transport solves the problem of transporting $\mu$ to $\nu$ while minimizing the cost function $c$. 
\rebuttal{In this work, we perform UDA between the source and target domains by using optimal transport to align the distributions of the source (space $\mathcal{X}$) and target (space $\mathcal{Y}$) intermediate network features.
The original Sinkhorn-Knopp's algorithm, which is a differentiable approximation of the Hungarian algorithm, is often employed to solve the optimal transport problem. However, it is subject to the mass preservation constraint, \textit{i.e.}, every feature $ x \in \mathcal{X} $ has to be matched to one or more features in $\mathcal{Y}$. This is often unrealistic when dealing with dissimilar scenes such as those from a highway and an urban city.
We address this limitation by redefining the problem as an unbalanced optimal transportation formulation which relaxes this constraint and allows for the destruction of mass between the two domains. In other words, this relaxation allows for features to not be matched between the source and target domains, thus making optimal transport robust to dissimilar scenes. We implement unbalanced optimal transport using the extension of the Sinkhorn-Knopp's algorithm proposed in~\cite{chizat2018scaling} which approximates it in a fast, highly parallelizable, and differentiable manner.} This formulation thus allows for the creation of a loss function that is robust to both stochasticity induced by minibatch sampling as well as undesired coupling between the samples~\cite{pmlr-v139-fatras21a}.
We define the cost function $ c_l $ as a function between $ i $-th source feature vector $ \bm{\psi}^{(s)}_{l, i} $ and $ j $-th target feature vector $ \bm{\psi}^{(t)}_{l, j} $ at layer $ l $ as\looseness=-1
\begin{equation}
    c_l\left( i, j \right) = \left\Vert \bm{\psi}^{(s)}_{l, i} - \bm{\psi}^{(t)}_{l, j} \right\Vert_{2}^{2} + 
             \left\Vert \bm{\hat{y}}^{(s)}_{i} - \bm{\hat{y}}^{(t)}_{j} \right\Vert_{2}^{2},
\end{equation}
where $ \bm{\hat{y}}^{(s)}_{i} $ and $ \bm{\hat{y}}^{(t)}_{j} $ are the source and target output features before the classification layer respectively. We solve the unbalanced optimal transport and perform domain adaptation on multiple scales $l \in \{l_1, l_2, \cdotp\mkern-2mu\cdotp\mkern-2mu\cdotp, l_n\}$ of intermediate network features, resulting in a set of transport plans $ \bm{\pi^{*}}_{l}  $.
We name this approach \textit{Multi-Scale Feature-Space Optimal Transport}~(MS-FSOT).

We optimize the model by alternating between solving the optimal transport and the deep neural network gradient descent step. After optimal transport is solved for transport plans $ \bm{\pi^{*}_{l}} $, we optimize the panoptic segmentation network by minimizing the following loss function:
\begin{equation}
    \mathcal{L} = \mathcal{L}_{\,\text{task}}\left(\bm{\mathcal{X}}^{(s)}, \bm{y}^{(s)}\right) + \sum_{l} \sum_{i,j} \bm{\pi^{*}}_{l,ij}  c_l(i, j),
\end{equation}
where $ \bm{y}^{(s)} $ are source data panoptic labels and $ \mathcal{L}_{\,\text{task}} $ is any panoptic loss function that prevents the degradation of model performance in the source domain.

\subsubsection{Instance-Aware Sampling (IAS)}

Existing approaches to solve optimal transport do not scale sufficiently to meet the requirements of existing computer vision tasks.
This limitation is typically addressed by employing a stochastic approximation using mini-batches from both the source and target domains~\cite{Damodaran_2018_ECCV}. However, solving optimal transport differentiably and iteratively on multiple scales is still not feasible due to the space complexity being quadratic in the number of features. One way to address this issue is to sub-sample the features for optimal transport beforehand. However, random sub-sampling of features mostly captures global statistics which causes a problem in scenes with severe class and object-size imbalance. Specifically, with significant intra-class variance of instance sizes, instances occupying a large number of pixels are over-represented while smaller instances are under-represented, which subsequently suppresses the local geometrical structures in the feature space. We also observe a similar behavior when few \textit{stuff} classes dominate the scene.

To overcome the aforementioned problems, we propose an \textit{Instance-Aware Sampling} (IAS) strategy that leverages complete panoptic information, \textit{i.e.}, both class and instance labels.
Specifically, we sample a fixed number of points, $64$ in our setting, for each $ \left(\text{class}, \text{instance}\right) $ pair for alignment by MS-FSOT.
In this way, we capture global statistics while preserving the local information of every class and instance, which is crucial for an optimal alignment of the source and target domains. Further, we downsample the panoptic labels to match the height and width of the $l$\textsuperscript{th} feature map to associate every element in the feature map with its corresponding $ \left(\text{class}, \text{instance}\right) $ pair. Particularly, we use the groundtruth panoptic labels for the source domain, while employing the panoptic predictions from the network for the target domain. Since the panoptic predictions are inaccurate at the beginning of the training, we use a linear curriculum learning schedule wherein we gradually transition from random sampling to IAS over the course of training. Additionally, we prevent matches between \textit{stuff} and \textit{thing} categories by constraining optimal transport by setting the corresponding entries in $ \bm{\pi^{*}_l} $ to zero and re-normalizing it.
\vspace{-2mm}

\subsection{Inference}
\label{sec:inference}

Before evaluation on target data, we use an efficient formulation of progressive domain calibration (PDC)~\cite{wu2018squeezesegv2} to re-calibrate the batch normalization parameters of the adapted model.
Batch normalization parameters typically accumulate source data-specific characteristics which preserve a certain domain gap between the source and target domains.
To minimize this domain gap, PDC recalculates the mean and variance for every batch normalization layer on the target data. However, since these statistics are progressively computed on each batch normalization layer, PDC incurs a significant time penalty. For instance, on the nuScenes dataset containing $34149$ samples, PDC requires nearly \SI{810}{\sec} to compute the batch normalization statistics. We address this high computational requirement by proposing an efficient version of PDC, called PDC-Lite. PDC-Lite is based on the observation that re-calibrating just the first batch normalization layer is sufficient since the following batch normalization layers operate on previously normalized feature activations. 
Our experiments show that PDC-Lite is more than $26$-times faster than PDC, computing the batch norm statistics on the nuScenes dataset in only \SI{30}{\sec} while achieving similar performance.
\rebuttal{It is important to note that the PDC-lite is a one-time operation performed before deploying the model, and thus does not influence the actual inference time of the adapted model}.
\looseness=-1
\section{Experimental Results}
\label{sec:experiments}

In this section, we present quantitative and qualitative
results of our proposed \net\ model, and a detailed ablation study to establish the utility of our contributions. We also describe the datasets that we employ and the training protocol.
\rebuttal{We perform all the experiments using an Intel Core i5-6500 CPU and an NVIDIA Titan X GPU.}
\vspace{-1mm}

\subsection{Datasets}
\label{subsec:datasets}
We empirically evaluate our domain adaptation approach on three dataset pairs, namely, (i) SemanticKITTI~\cite{behley2019iccv} $\rightarrow$ nuScenes~\cite{nuscenes2019}, (ii) SemanticKITTI~\cite{behley2019iccv} $\rightarrow$ PandaSet~\cite{pandaset}, and (iii) nuScenes~\cite{nuscenes2019} $\rightarrow$ PandaSet~\cite{pandaset}. \rebuttal{We choose these dataset pairs to account for diverse real-world scenarios such as different number of LiDAR scan lines (i, iii) and different vertical scan line patterns (ii, iii).} 
\acceptance{Specifically, SemanticKITTI and PandaSet have 64 scan lines each, while nuScenes has only 32 scan lines. Further, SemanticKITTI and nuScenes contain a uniform vertical distribution of LiDAR scan lines, while PandaSet has a non-uniform scan line distribution which is denser around the center of its vertical field-of-view.}
To accommodate the input requirements of the base panoptic models, we project the 3D LiDAR points onto a 2D image having a fixed width of \SI{1024}{px} and height equaling to the number of scan lines of the LiDAR. Accordingly, SemanticKITTI and PandaSet have an input image resolution of $1024 \times 64$ pixels while nuScenes has a resolution of $1024 \times 32$ pixels.
To allow comparison between datasets containing varying numbers of semantic classes, we map\footnote{Label mappings: \url{https://github.com/robot-learning-freiburg/AdaptLPS}} the semantic labels in SemanticKITTI and PandaSet to those defined in nuScenes. We divide the datasets into training and validation sets following the splits defined in~\cite{behley2019iccv} and \cite{cit:panoptic-nuscenes} for SemanticKITTI and nuScenes respectively. For PandaSet, we numerically sort all sequences and use the last $20 \%$ sequences for validation.

\subsection{Training Setup}
\label{subsec:training-setting}

We demonstrate the effectiveness of our UDA approach by incorporating \net\ into two state-of-the-art panoptic segmentation networks, namely, EfficientLPS~\cite{sirohi2021efficientlps} and Milioto~\textit{et.~al.}~\cite{milioto2020lidar}.
\rebuttal{Although both networks are projection-based, our AdaptLPS is independent of the architecture and can be applied to any types of LiDAR panoptic segmentation approach, with the only required modification being the IMN architecture which should be adapted.}
We first pre-train both networks on the source domain and subsequently adapt them to the target domain by fine-tuning the network with our domain adaptation strategy on both the source and target domains simultaneously.
We train EfficientLPS~\cite{sirohi2021efficientlps} using SGD with an initial learning rate (LR) of $0.07$, a momentum of $0.9$, and an LR schedule that decays LR by a factor of $10$ after $16000$ and $22000$ iterations. We use layers $ l \in \left\{ \text{RP}_{4}, \text{RP}_{8}, \text{RP}_{16}, \text{RP}_{32} \right\} $ as input to MS-FSOT. For BonnetalPS, we use the Adam optimizer with an initial LR of $0.0001$, first and second momenta of $\left(0.9, 0.99\right)$ respectively, and an LR schedule that multiplies LR by $0.99$ after each epoch. For this model, we align layers $ l \in \left\{ \text{OS}_{4}, \text{OS}_{8}, \text{OS}_{16}, \text{OS}_{32} \right\} $ in MS-FSOT. We augment the training data with random horizontal flips and random horizontal crops of size \SI{768}{px} to prevent overfitting. We train each model with a batch size of $12$ for $200$ epochs and employ early stopping with a patience of $5$ epochs on the panoptic quality (PQ) metric to further mitigate model overfitting.\looseness=-1

\subsection{Quantitative Results}
\label{subsec:sota-comparison}

\begin{table}
\setlength{\tabcolsep}{5pt}
\footnotesize
\centering
\caption{Cross-domain panoptic segmentation performance. Note that SemanticKITTI and PandaSet labels are remapped to that of nuScenes to facilitate comparison (refer \secref{subsec:datasets}).}
\begin{tabular}{cc|cc}
\toprule
\textbf{Source Dataset} & \textbf{Target Dataset} & \multicolumn{2}{c}{\textbf{PQ Score}} \\
& & \textbf{EfficientLPS~\cite{sirohi2021efficientlps}} & \rebuttal{\textbf{BonnetalPS~\cite{milioto2020lidar}}} \\
\midrule
    SemanticKITTI & SemanticKITTI & 48.28 & \rebuttal{39.57} \\
    SemanticKITTI & nuScenes & 10.24 & \rebuttal{8.16} \\
    SemanticKITTI & PandaSet & 9.64 & \rebuttal{7.67} \\
    \midrule
    nuScenes & nuScenes & 49.87 & \rebuttal{34.53} \\
    nuScenes & PandaSet & 18.73 & \rebuttal{15.67} \\
\bottomrule
\end{tabular}
\label{tab:source-target-pq}
\vspace{-3mm}
\end{table}

% Combined Table
\begin{table*}
    \footnotesize
    \centering
    \caption{Comparison of unsupervised domain adaptation methods for panoptic segmentation on three different pairs of datasets. Note that SemanticKITTI and PandaSet labels are remapped to that of nuScenes to facilitate comparison (refer \secref{subsec:datasets}).}
    \begin{tabular}{ccc|ccc|cccccc}
    \toprule
    \textbf{Datasets} & \textbf{Base Model} & \textbf{Domain Adaptation} & \textbf{PQ} & \textbf{SQ} & \textbf{RQ} & \textbf{PQ\textsuperscript{Th}} & \textbf{SQ\textsuperscript{Th}} & \textbf{RQ\textsuperscript{Th}} & \textbf{PQ\textsuperscript{St}} & \textbf{SQ\textsuperscript{St}} & \textbf{RQ\textsuperscript{St}} \\
        
        \midrule
        
        \parbox{6mm}{\multirow{10}{*}{\rotatebox[origin=c]{90}{\parbox{20mm}{\centering SemanticKITTI \\$\downarrow$ \\nuScenes}}}} & \multirow{4.5}{*}{EfficientLPS~\cite{sirohi2021efficientlps}} &  No Domain Adaptation & 10.24 & 36.16 & 15.96 & 8.68 & 35.19 & 14.28 & 12.20 & 36.87 & 18.82 \\
        & & GCA~\cite{morerio2018minimalentropy} & 18.17 & 46.93 & 26.55 & 16.08 & 45.61 & 23.40 & 21.97 & 47.95 & 31.59 \\
        & & UDABP~\cite{10.5555/3045118.3045244} & 18.59 & 47.08 & 27.04 & 16.00 & 45.83 & 24.15 & 22.28 & 47.61 & 32.34 \\
        \cmidrule{3-12}
        & & \net\ (Ours) & \textbf{25.00} & \textbf{48.84} & \textbf{29.23} & \textbf{21.68} & \textbf{47.63} & \textbf{25.98} & \textbf{30.50} & \textbf{49.71} & \textbf{34.53} \\
    
        \cmidrule{2-12}
        
        & \multirow{4.5}{*}{BonnetalPS~\cite{milioto2020lidar}} & No Domain Adaptation & 8.16 & 28.96 & 11.44 & 7.19 & 28.21 & 10.47 & 10.02 & 29.22 & 13.62 \\
        & & GCA~\cite{morerio2018minimalentropy} & 17.79 & 47.02 & 26.63 & 15.87 & 45.62 & 24.17 & 21.80 & 47.70 & 31.61 \\
        & & UDABP~\cite{10.5555/3045118.3045244} & 17.94 & 46.95 & 26.87 & 15.83 & 45.61 & 23.66 & 21.55 & 48.18 & 31.71 \\
        \cmidrule{3-12}
        & & \net\ (Ours) & \textbf{21.21} & \textbf{48.12} & \textbf{27.68} & \textbf{18.46} & \textbf{46.53} & \textbf{24.48} & \textbf{25.77} & \textbf{49.30} & \textbf{32.84} \\
        
        \midrule
        \midrule
        
        \parbox{6mm}{\multirow{10}{*}{\rotatebox[origin=c]{90}{\parbox{20mm}{\centering SemanticKITTI \\$\downarrow$ \\PandaSet}}}} & \multirow{4.5}{*}{EfficientLPS~\cite{sirohi2021efficientlps}} & No Domain Adaptation & 9.64 & 35.57 & 13.77 & 8.54 & 34.63 & 12.20 & 11.73 & 36.20 & 16.09 \\
        & & GCA~\cite{morerio2018minimalentropy} & 17.74 & 47.03 & 26.36 & 15.61 & 45.71 & 23.07 & 21.25 & 47.96 & 31.46 \\
        & & UDABP~\cite{10.5555/3045118.3045244} & 17.97 & 46.98 & 26.97 & 16.18 & 46.00 & 23.87 & 21.86 & 47.76 & 31.81 \\
        \cmidrule{3-12}
        & & \net\ (Ours) & \textbf{20.33} & \textbf{47.55} & \textbf{29.67} & \textbf{17.53} & \textbf{46.35} & \textbf{26.52} & \textbf{24.52} & \textbf{48.42} & \textbf{35.07} \\
        
        \cmidrule{2-12}
        
        & \multirow{4.5}{*}{BonnetalPS~\cite{milioto2020lidar}} & No Domain Adaptation & 7.67 & 28.89 & 10.29 & 6.97 & 27.85 & 9.20 & 9.65 & 29.31 & 12.03 \\
        & & GCA~\cite{morerio2018minimalentropy} & 15.45 & 46.86 & 24.49 & 13.47 & 45.82 & 21.76 & 18.97 & 47.55 & 29.48 \\
        & & UDABP~\cite{10.5555/3045118.3045244} & 15.98 & 47.00 & 25.81 & 13.92 & 45.81 & 22.76 & 19.76 & 47.87 & 31.00 \\
        \cmidrule{3-12}
        & & \net\ (Ours) & \textbf{19.19} & \textbf{47.91} & \textbf{27.80} & \textbf{16.83} & \textbf{46.87} & \textbf{24.99} & \textbf{23.29} & \textbf{48.68} & \textbf{32.99} \\
        
        \midrule
        \midrule
        
        \parbox{6mm}{\multirow{10}{*}{\rotatebox[origin=c]{90}{\parbox{20mm}{\centering nuScenes \\$\downarrow$ \\PandaSet}}}} & \multirow{4.5}{*}{EfficientLPS~\cite{sirohi2021efficientlps}} & No Domain Adaptation & 18.73 & 35.30 & 20.53 & 16.58 & 34.85 & 17.98 & 22.19 & 35.90 & 24.38 \\
        & & GCA~\cite{morerio2018minimalentropy} & 25.90 & 48.06 & 32.84 & 22.45 & 46.61 & 28.98 & 31.21 & 48.85 & 38.88 \\
        & & UDABP~\cite{10.5555/3045118.3045244} & 25.01 & 47.95 & 32.06 & 21.70 & 46.72 & 28.51 & 30.21 & 48.56 & 37.65 \\
        \cmidrule{3-12}
        & & \net\ (Ours) & \textbf{27.76} & \textbf{48.86} & \textbf{34.37} & \textbf{24.18} & \textbf{47.61} & \textbf{30.45} & \textbf{33.54} & \textbf{49.47} & \textbf{40.65} \\
        
        \cmidrule{2-12}
        
        & \multirow{4.5}{*}{BonnetalPS~\cite{milioto2020lidar}} & No Domain Adaptation & 15.67 & 30.04 & 18.40 & 13.48 & 29.28 & 16.37 & 18.80 & 30.47 & 21.93 \\
        & & GCA~\cite{morerio2018minimalentropy} & 23.66 & \textbf{48.62} & 30.58 & 20.49 & 47.03 & 27.09 & 28.44 & 49.30 & 36.14 \\
        & & UDABP~\cite{10.5555/3045118.3045244} & 24.83 & 48.50 & 31.22 & 21.59 & 46.85 & 27.57 & 29.56 & 49.19 & 37.02 \\
        \cmidrule{3-12}
        & & \net\ (Ours) & \textbf{26.19} & 48.46 & \textbf{33.19} & \textbf{22.78} & \textbf{47.22} & \textbf{29.80} & \textbf{31.58} & \textbf{49.57} & \textbf{39.50} \\
    
        \bottomrule
    \end{tabular}
    \label{tab:panoptic-da}
    \vspace{-3mm}
    \end{table*}

Since we are the first to tackle the problem of LiDAR panoptic domain adaptation, we evaluate the performance of our \net\ strategy by comparing it with four novel baselines. We create the baselines by combining two state-of-the-art panoptic segmentation approaches, EfficientLPS~\cite{sirohi2021efficientlps} and Milioto~\textit{et~al.}~\cite{milioto2020lidar}, with two recent UDA methods, namely, Geodesic Correlation Alignment (GCA)~\cite{morerio2018minimalentropy} and UDA by backpropagation (UDABP)~\cite{10.5555/3045118.3045244}. For brevity, we name the model proposed by Milioto~\textit{et~al.} as BonnetalPS. We report the performance using the panoptic metrics of panoptic quality (PQ), segmentation quality (SQ), and recognition quality (RQ) \rebuttal{as defined in~\cite{mohan2020efficientps}. We independently calculate the PQ score for each class before averaging it over all classes.}
We also separately report the aforementioned metrics for the \textit{stuff} and \textit{thing} classes for the sake of completeness. 

Before evaluating \net, we highlight the need for domain adaptation by quantifying the performance drop induced by the domain gap between the source and target datasets. \tabref{tab:source-target-pq} presents the PQ values obtained by EfficientLPS when it is trained and evaluated on different datasets. We observe that the PQ value suffers a significant drop when the target dataset differs from the source dataset. For instance, EfficientLPS trained and evaluated on SemanticKITTI achieves a PQ score of $48.28\%$, which drops to $10.24\%$ and $9.64\%$ when evaluated on nuScenes and PandaSet respectively. We make a similar observation when EfficientLPS is trained on nuScenes. 
This significant drop in performance demonstrates the need to employ domain adaptation to improve the cross-domain performance of the model.

% KITTI -->  nuScenes Semantic Domain Adaptation
\begin{table*}
\footnotesize
\centering
\caption{Comparison of unsupervised domain adaptation methods for semantic segmentation on the SemanticKITTI $ \rightarrow $ nuScenes dataset pair.}
\setlength\tabcolsep{3.7pt}
 \begin{tabular}{c|ccccccccccccc|c}
 \toprule
  \textbf{Method} & \textbf{Road} & \textbf{Side.} & \textbf{Build.} & \textbf{Fence} & \textbf{Pole} & \textbf{Sign} & \textbf{Veg.} & \textbf{Terrain} & \textbf{Trunk} & \textbf{Person} & \textbf{Car} & \textbf{Oth.Veh.} & \textbf{Bicycle} & \textbf{mIoU}
 
 \\
 \midrule

CP + GCA~\cite{langer2020iros} & 88.5 & 53.0 & 80.9 & 43.0 & 35.3 & 16.3 & 41.6 & 0.8 & 3.4 & 18.6 & 70.8 & \textbf{6.4} & 8.8 & 35.9 \\ 
MB + GCA~\cite{langer2020iros} & 88.9 & 45.2 & 79.0 & 31.2 & 33.7 & 14.1 & 38.3 & 1.2 & 2.9 & 15.3 & 63.7 & 4.8 & 5.8 & 32.6 \\ 
MC + GCA~\cite{langer2020iros} & 89.3 & 49.8 & 80.0 & 42.1 & 34.2 & 17.3 & 44.7 & 2.9 & 3.9 & 20.9 & 65.2 & 3.1 & 6.7 & 35.4 \\

\midrule

EfficientLPS~\cite{sirohi2021efficientlps} + \net\ (Ours) & \textbf{89.7} & 55.0 & \textbf{81.4} & \textbf{50.7} & \textbf{35.5} & \textbf{19.1} & \textbf{48.3} & \textbf{4.0} & \textbf{6.5} & \textbf{23.0} & \textbf{71.5} & 6.0 & \textbf{9.5} & \textbf{38.5} \\
BonnetalPS~\cite{milioto2020lidar} + \net\ (Ours) & 88.1 & \textbf{56.0} & 81.3 & 49.8 & \textbf{35.5} & 18.4 & 45.9 & 1.8 & 3.1 & 22.0 & 71.4 & 4.7 & 9.0 & 37.5 \\

 \bottomrule
 \end{tabular}
\label{tab:kitti-nuscenes-da-semseg}
\vspace{-3mm}
\end{table*}

\tabref{tab:panoptic-da} compares the performance of \net\ with existing domain adaption strategies on three dataset pairs. For brevity, we describe the quantitative results in the following text using only EfficientLPS as the base model. Nevertheless, a similar argument can also be extended to the approaches using BonnetalPS as the base model. We observe that our proposed \net\ domain adaptation strategy outperforms all the baselines on all three dataset pairs. For SemanticKITTI $\rightarrow$ nuScenes, \net\ exceeds the best baseline, UDABP, by \SI{6.41}{pp} where a majority of the improvement comes from the \SI{5.68}{pp} increase in PQ\textsuperscript{Th} score. This improvement in the PQ score can be attributed to both IMN and our unbalanced optimal transport formulation which minimizes the gap between the different intensity models and efficiently handles dissimilar features between the two domains. Further, the \textit{thing} classes benefit from IAS which enables the detection of finer instance-level features and subsequently helps in the detection of more instances in the target domain. We make a similar observation for SemanticKITTI $\rightarrow$ PandaSet, where a \SI{2.36}{pp} improvement in the PQ score compared to UDABP is largely contributed by a \SI{1.35}{pp} increase in the PQ\textsuperscript{Th} score. However, the improvement achieved by \net\ is less prominent for this dataset pair due to PandaSet being more than $5$-times smaller in size and much less diverse as compared to nuScenes, which limits the domain adaptation potential. For the nuScenes $\rightarrow$ PandaSet dataset pair, we observe that our model achieves a limited improvement over the best baseline, GCA, exceeding it by only \SI{1.86}{pp}. This is due to the fact that nuScenes and PandaSet contain the least domain gap among all the dataset pairs as reflected by the \rebuttal{higher} ``No Domain Adaptation'' PQ scores. \rebuttal{Moreover, both datasets use similar intensity calibrations which reduces the performance gain obtained from IMN resulting in a decrease in the margin by which our method outperforms other methods.} Consequently, the GCA algorithm effectively aligns the second-order statistics between nuScenes and PandaSet, and achieves a high PQ score.
Nevertheless, our \net\ strategy consistently outperforms both the domain adaptation baselines across all the dataset pairs, thus demonstrating its versatility and effective generalization ability.\looseness=-1

We further investigate the generalizability of our \net\ strategy by comparing it against UDA approaches proposed in~\cite{langer2020iros} on the semantic segmentation task. We employ modified variants of EfficientLPS~\cite{sirohi2021efficientlps} and BonnetalPS~\cite{milioto2020lidar} as the base networks, wherein we remove the instance head and panoptic fusion module to make it compatible with the semantic segmentation task. Further, we also adapt \net\ to the semantic segmentation task by replacing IAS with class-based sampling to account for the lack of knowledge of instances. We primarily compare the models using the mIoU metric but also present class-wise IoUs for completeness. To ensure a fair comparison with the baselines, we follow the evaluation protocol proposed in~\cite{milioto2020lidar}, i.e., we train the model on SemanticKITTI and evaluate it on the manually annotated nuScenes sequence \textit{scene-0103} containing $389$ LiDAR scans. \tabref{tab:kitti-nuscenes-da-semseg} presents the results of this experiment. We observe that our \net\ strategy outperforms the state-of-the-art approaches, exceeding them by \SI{2.6}{pp} and \SI{1.6}{pp} on the mIoU metric when using the base models of EfficientLPS and BonnetalPS respectively. Further, we observe that small classes such as fence, sign, person, and bicycle show a significant improvement in the IoU scores which demonstrates the effectiveness of class-based sampling in allowing an equal representation of small classes in the optimal transport cost matrix. Lastly, given that our \net\ strategy exceeds the baselines across both base models, we conclude that our domain adaption strategy is easily adaptable to existing semantic and panoptic segmentation approaches which allows for its widespread use in multiple applications.
\vspace{-1mm}

\begin{table*}
\centering
\footnotesize
\caption{Ablation study on different components of the \net 
~architecture.}
\setlength\tabcolsep{3.7pt}
 \begin{tabular}{c|ccccc|ccc|cccccc}
 \toprule
 \textbf{Model} & \textbf{PC} & \textbf{IMN} & \textbf{PDC-Lite} & \textbf{MS-FSOT} & \textbf{IAS} & \textbf{PQ} & \textbf{SQ} & \textbf{RQ} & \textbf{PQ\textsuperscript{Th}} & \textbf{SQ\textsuperscript{Th}} & \textbf{RQ\textsuperscript{Th}} & \textbf{PQ\textsuperscript{St}} & \textbf{SQ\textsuperscript{St}} & \textbf{RQ\textsuperscript{St}} \\
 \midrule

M1 & - & - & - & - & - & 10.27 & 36.12 & 15.94 & 8.90 & 35.08 & 14.04 & 12.52 & 36.56 & 18.88 \\
M2 & \checkmark & - & - & - & - & 11.97 & 38.55 & 16.67 & 10.41 & 37.22 & 14.81 & 14.63 & 39.23 & 19.89 \\
M3 & \checkmark & \checkmark & - & - & - & 13.47 & 40.51 & 18.11 & 11.56 & 39.31 & 16.16 & 16.47 & 41.38 & 21.39 \\
M4 & \checkmark & \checkmark & \checkmark & - & - & 14.25 & 40.22 & 19.19 & 12.38 & 38.97 & 17.18 & 17.21 & 40.89 & 22.77 \\
M5 & \checkmark & \checkmark & \checkmark & \checkmark & - & 23.19 & 48.35 & 28.69 & 20.57 & 47.02 & 25.61 & 27.84 & 49.05 & 33.96 \\
\midrule
M6 & \checkmark & \checkmark & \checkmark & \checkmark & \checkmark & \textbf{25.00} & \textbf{48.84} & \textbf{29.23} & \textbf{21.68} & \textbf{47.63} & \textbf{25.98} & \textbf{30.50} & \textbf{49.71} & \textbf{34.53} \\

 \bottomrule
 \end{tabular}
\label{tab:network-ablation}
\vspace{-3mm}
\end{table*}

\subsection{Ablation Study}
\label{subsec:ablation-study}

% Qualitative Results
\begin{figure*}
\centering
\footnotesize
\setlength{\tabcolsep}{0.05cm}% for the horiz padding
{
\renewcommand{\arraystretch}{0.15}% for the vertical padding
\newcolumntype{M}[1]{>{\centering\arraybackslash}m{#1}}
\begin{tabular}{M{0.4cm}M{3.8cm}M{3.8cm}M{3.8cm}M{3.8cm}}
& No Domain Adaptation & Best Baseline & \net\ (Ours) & Improvement/Error Map \\
\\
\\
\\
\rotatebox[origin=c]{90}{(a) K $\rightarrow $ N} & {\includegraphics[width=\linewidth, frame]{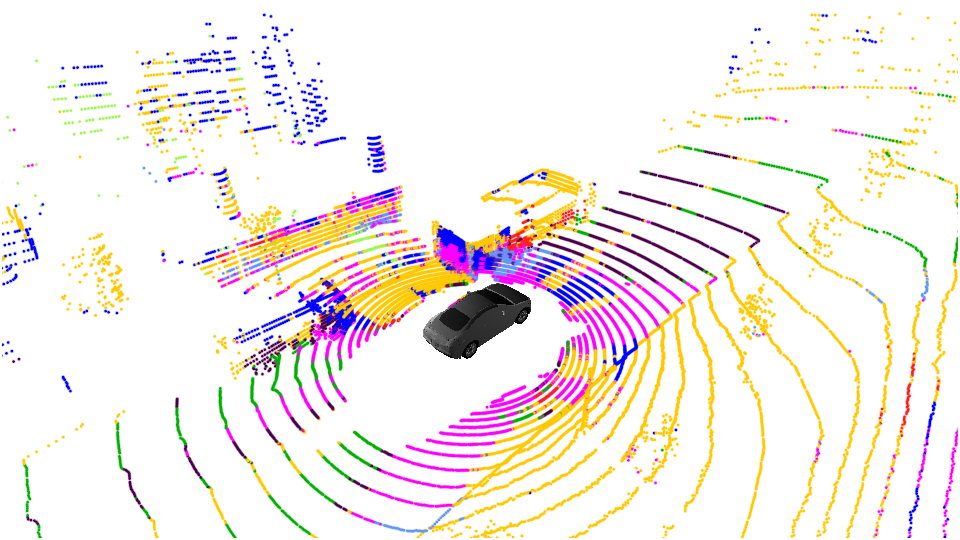}} & {\includegraphics[width=\linewidth, frame]{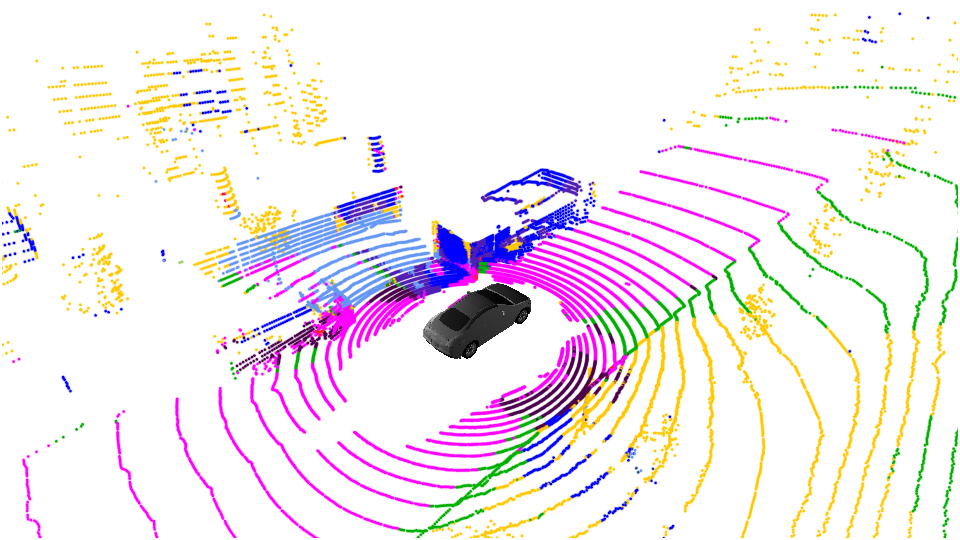}} & {\includegraphics[width=\linewidth, frame]{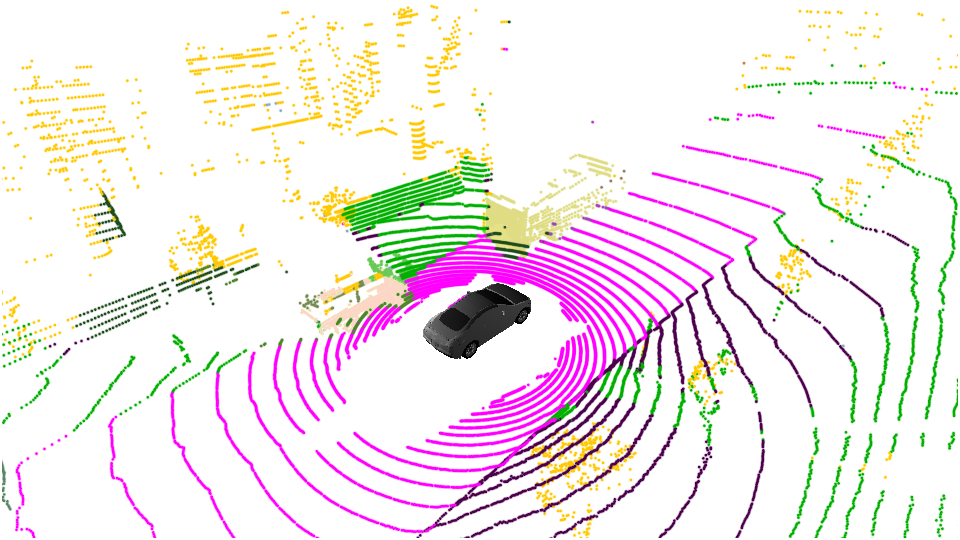}}
&
{\includegraphics[width=\linewidth, frame]{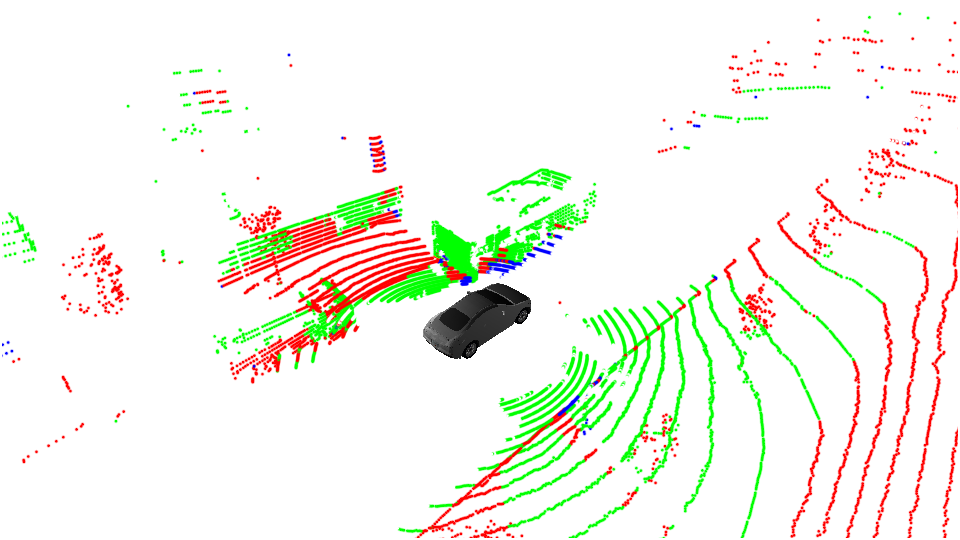}}
\\
\\
\rotatebox[origin=c]{90}{(b) K $\rightarrow $ N} & {\includegraphics[width=\linewidth, frame]{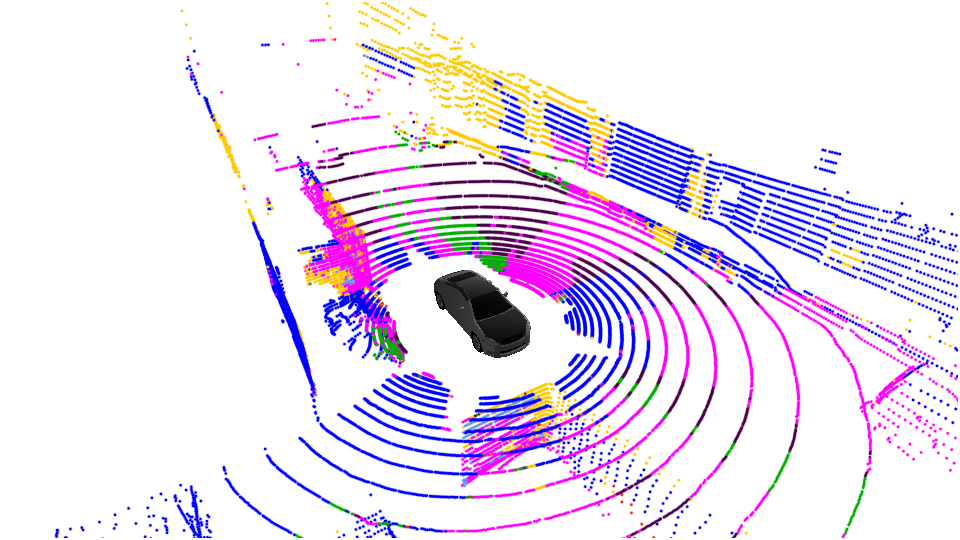}} & {\includegraphics[width=\linewidth, frame]{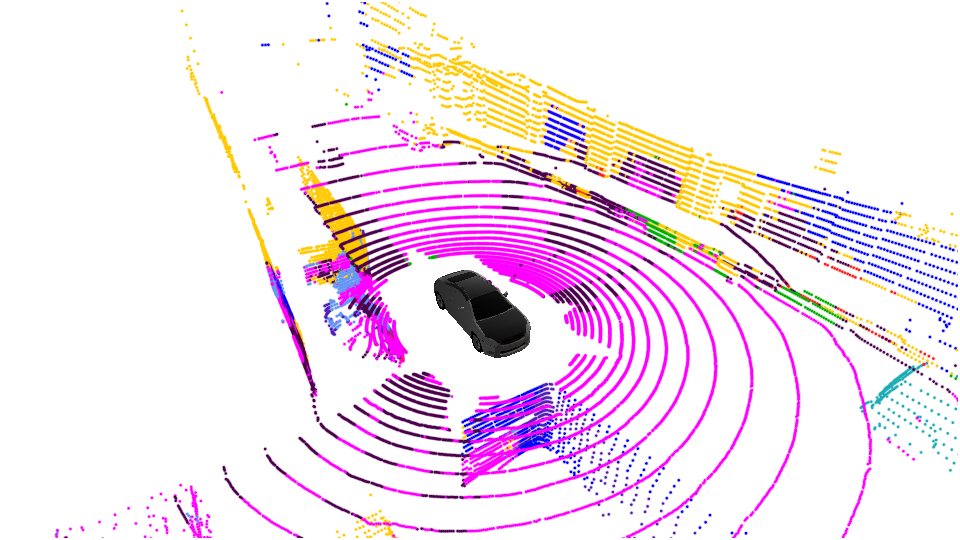}} & {\includegraphics[width=\linewidth, frame]{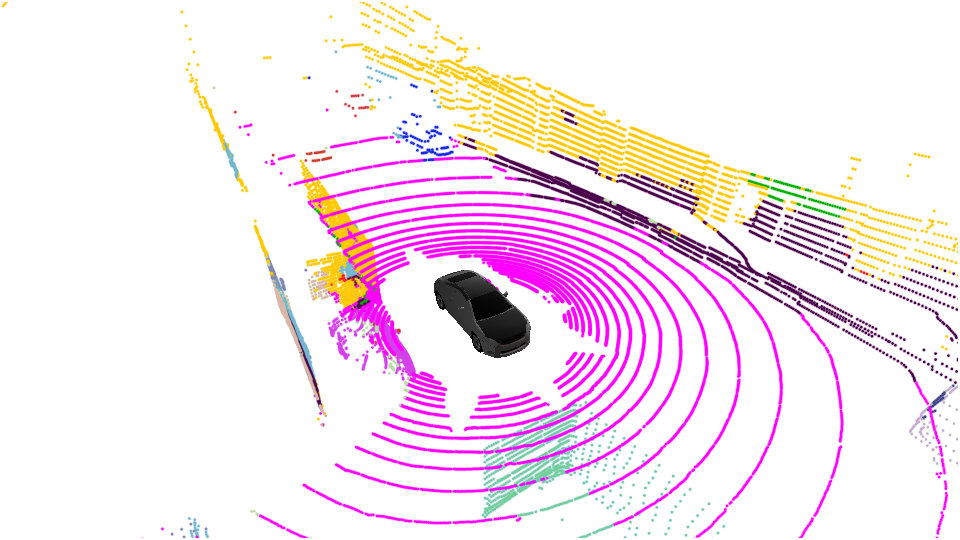}}
&
{\includegraphics[width=\linewidth, frame]{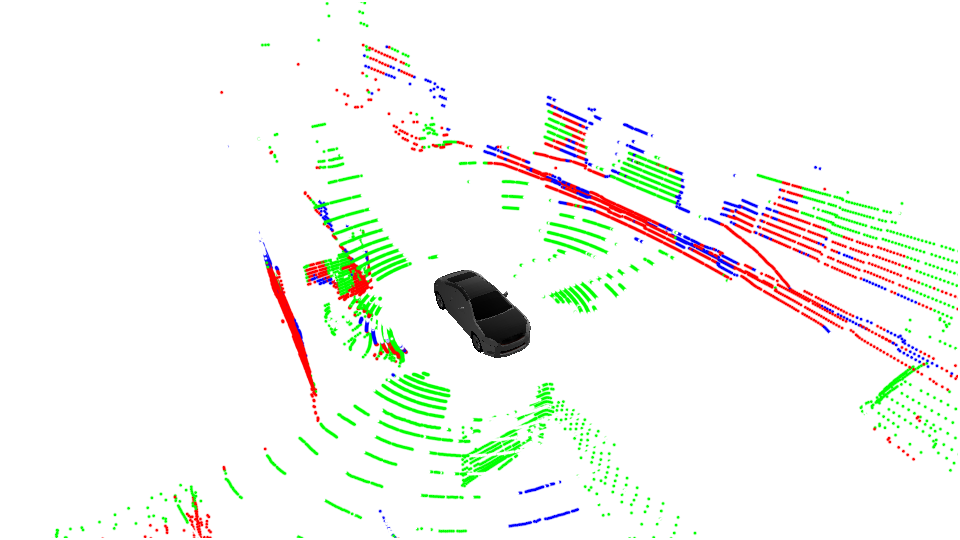}}
\\
\\
\rotatebox[origin=c]{90}{(c) K $ \rightarrow $ P} & {\includegraphics[width=\linewidth,, frame]{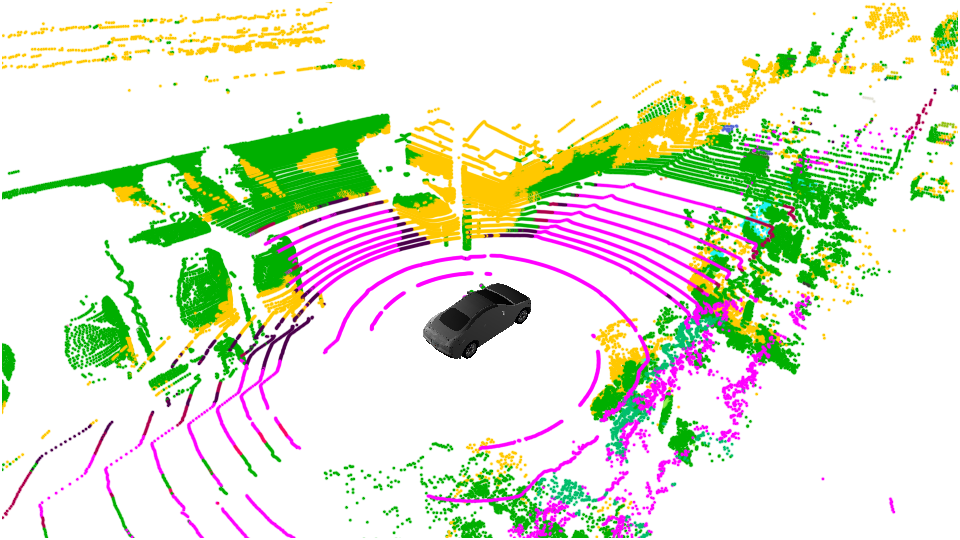}} & {\includegraphics[width=\linewidth, frame]{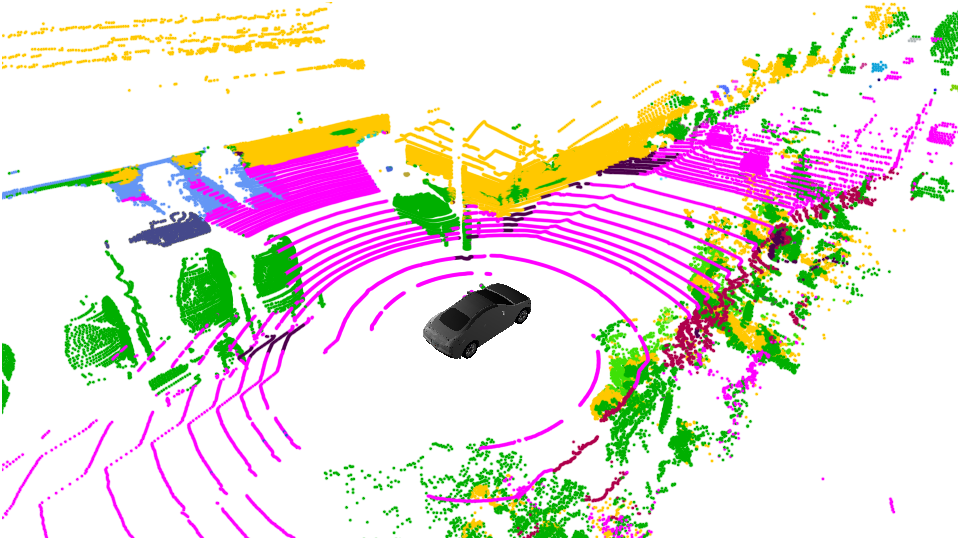}} & {\includegraphics[width=\linewidth, frame]{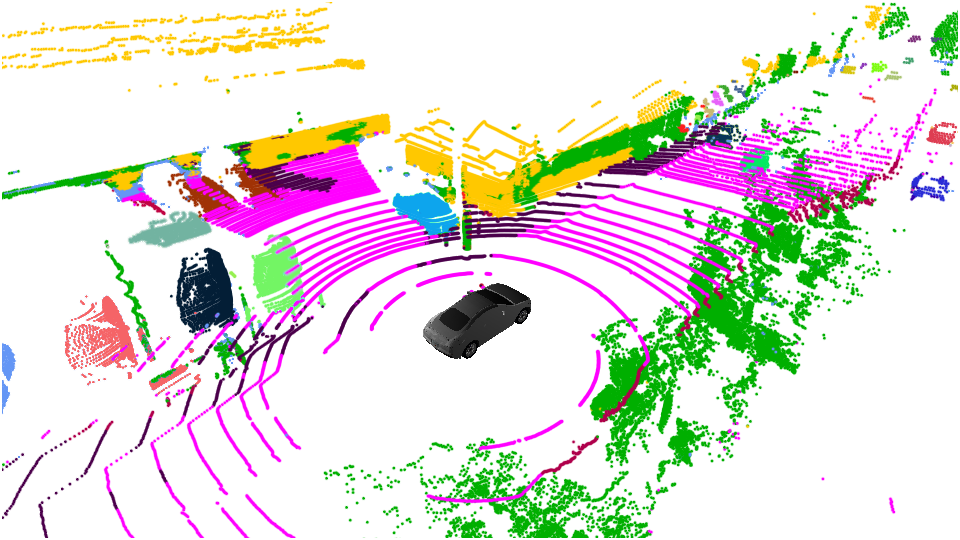}}
&
{\includegraphics[width=\linewidth, frame]{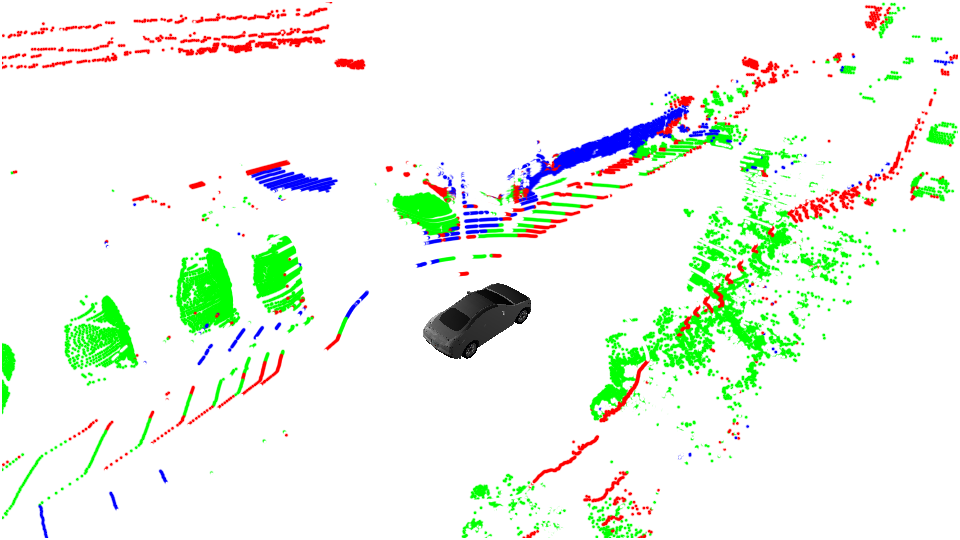}}
\\
\\
\rotatebox[origin=c]{90}{(d) K $ \rightarrow $ P} & {\includegraphics[width=\linewidth,, frame]{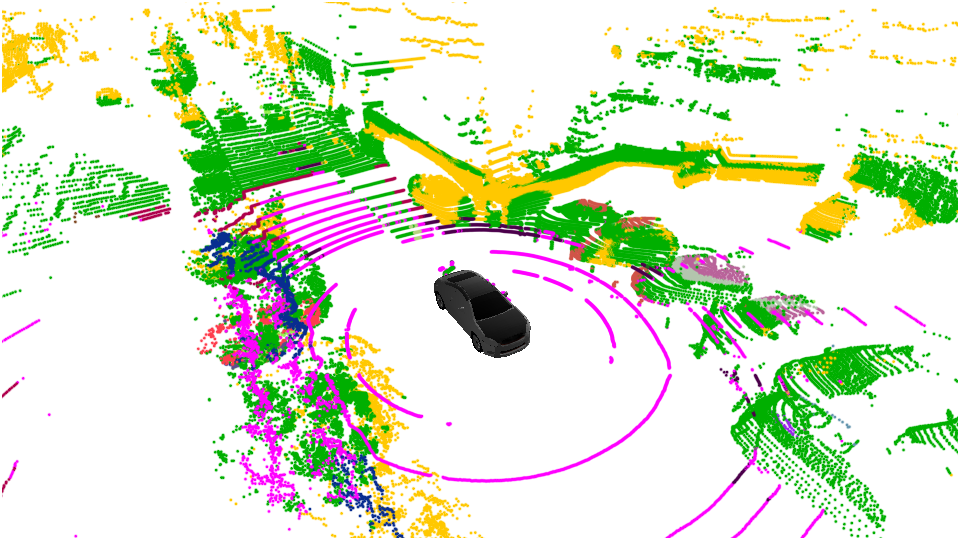}} & {\includegraphics[width=\linewidth, frame]{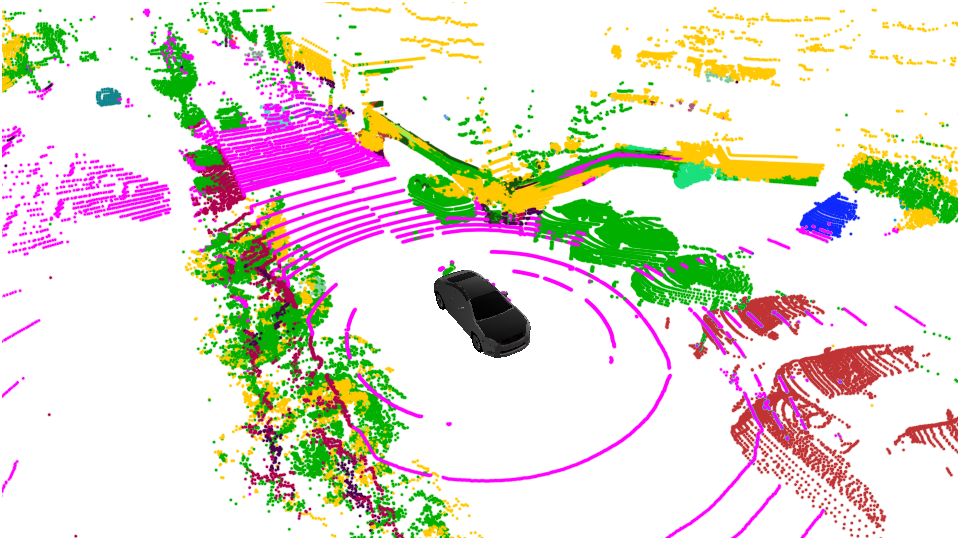}} & {\includegraphics[width=\linewidth, frame]{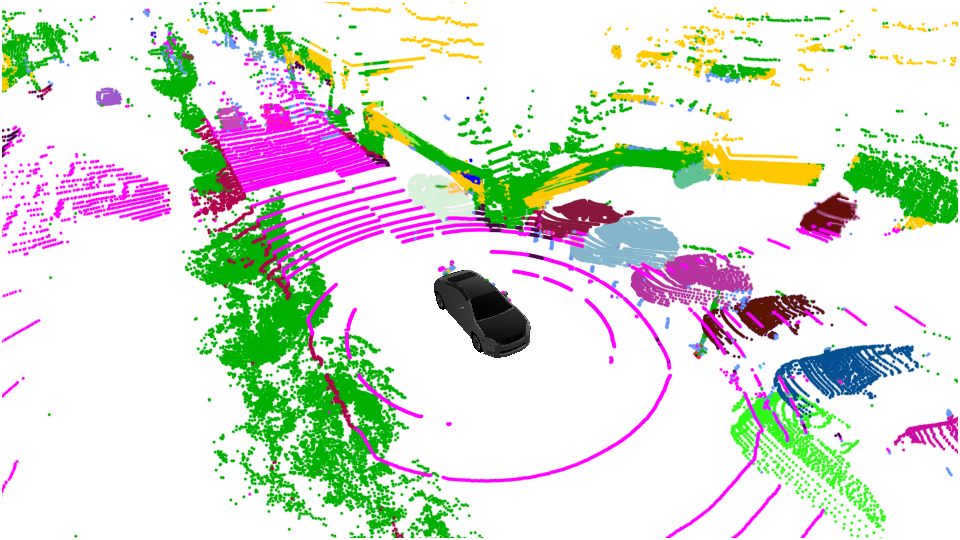}}
&
{\includegraphics[width=\linewidth, frame]{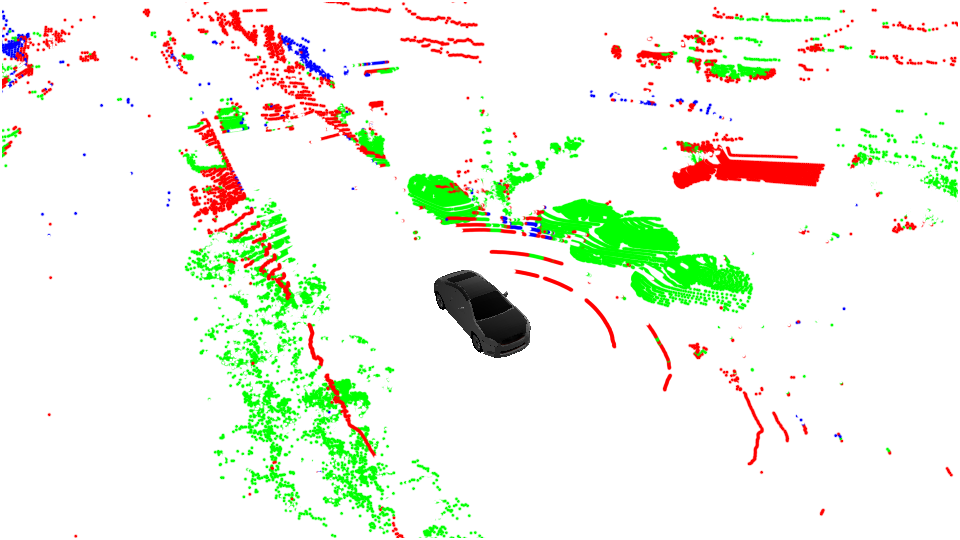}}
\\
\\
\rotatebox[origin=c]{90}{(e) N $ \rightarrow $ P} & {\includegraphics[width=\linewidth,, frame]{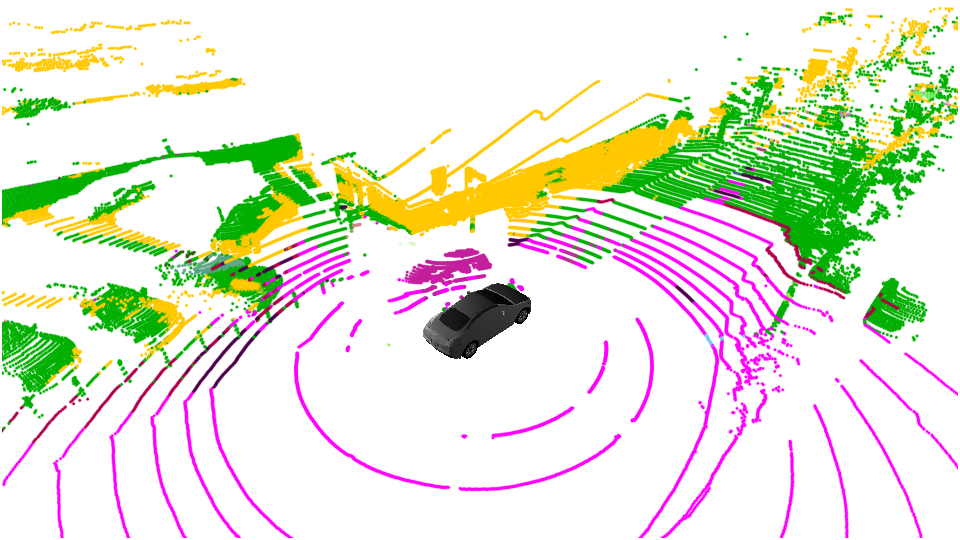}} & {\includegraphics[width=\linewidth, frame]{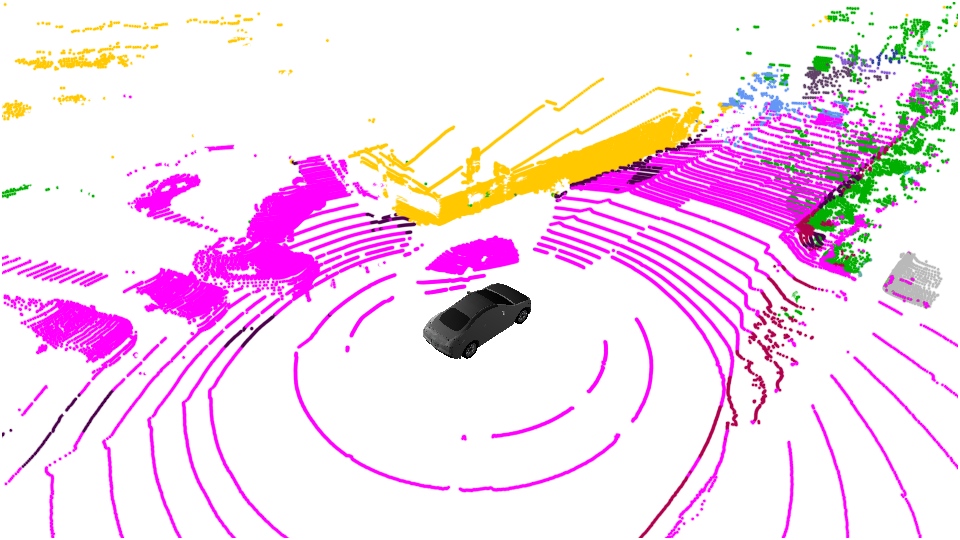}} & {\includegraphics[width=\linewidth, frame]{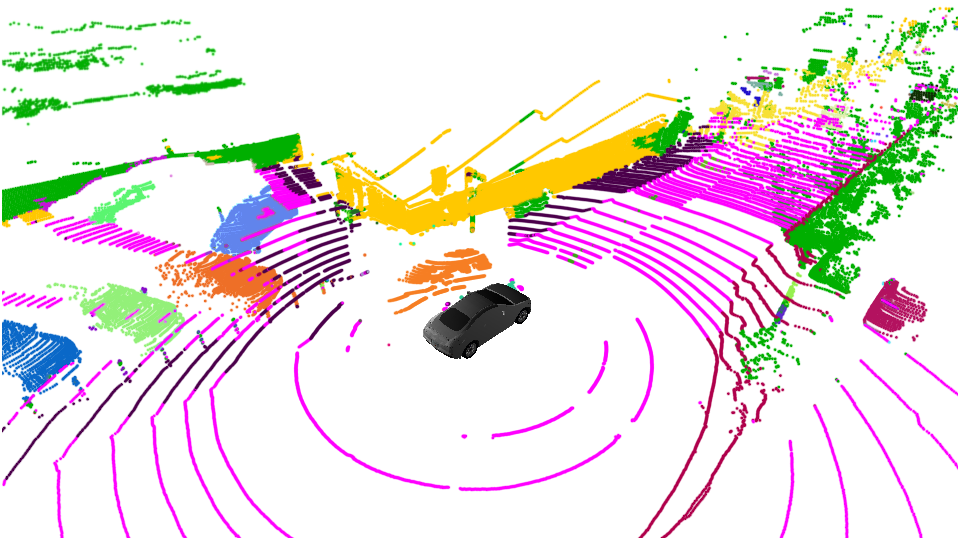}}
&
{\includegraphics[width=\linewidth, frame]{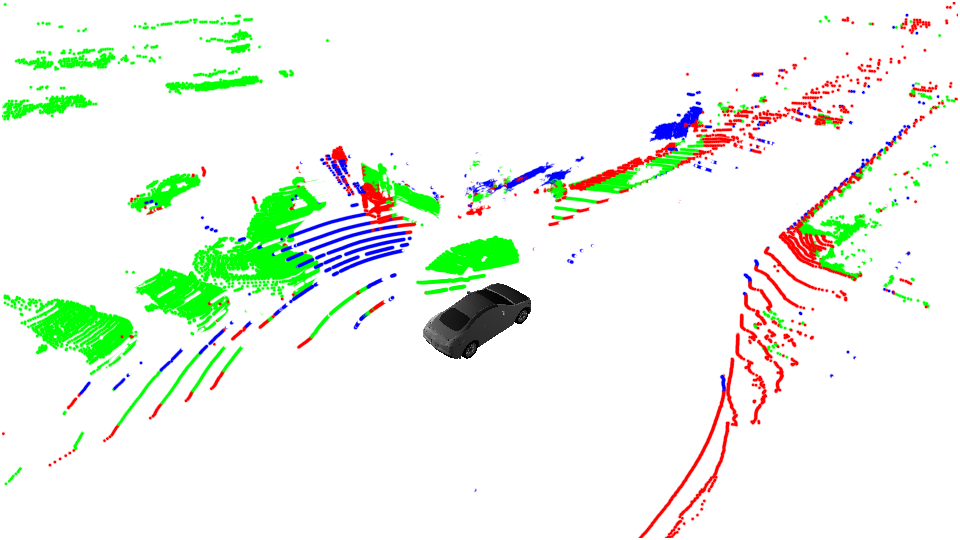}}
\\
\\
\rotatebox[origin=c]{90}{(f) N $ \rightarrow $ P} & {\includegraphics[width=\linewidth,, frame]{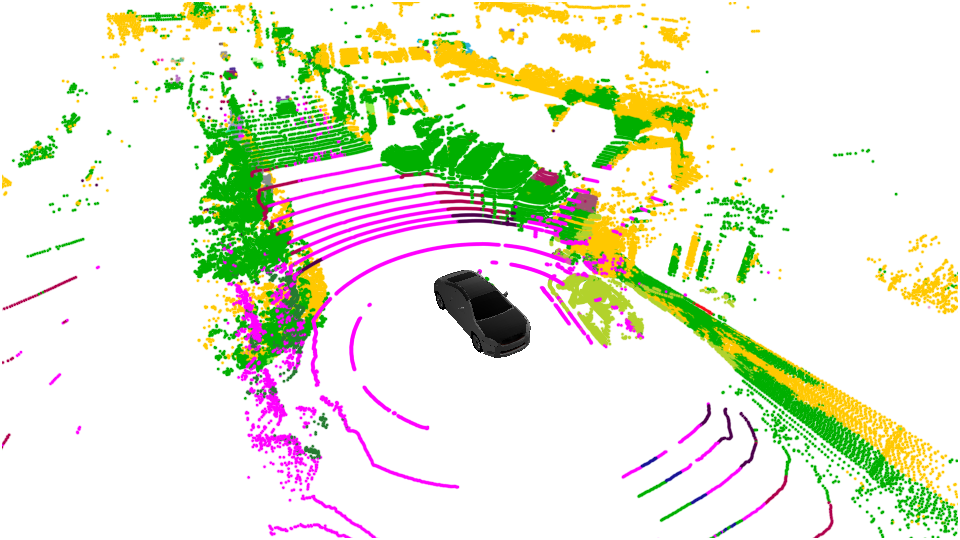}} & {\includegraphics[width=\linewidth, frame]{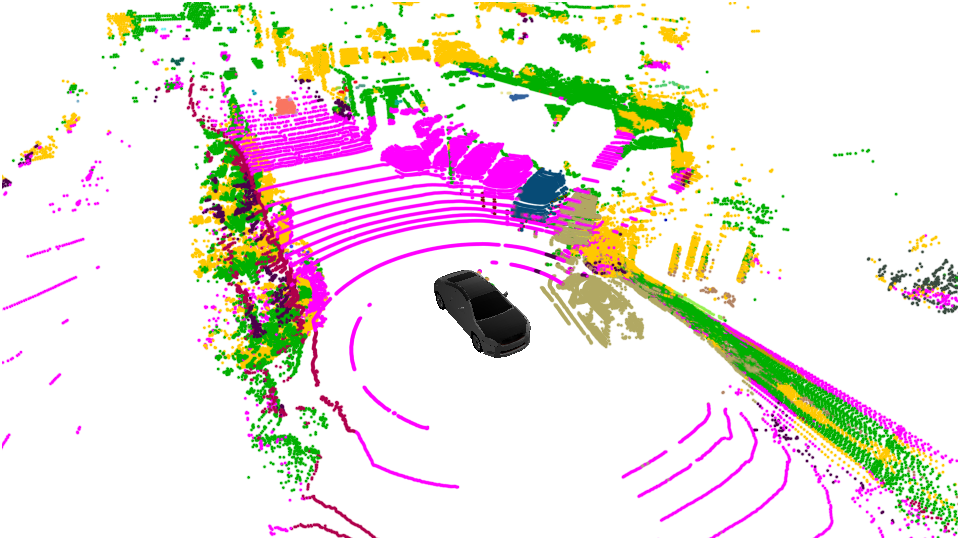}} & {\includegraphics[width=\linewidth, frame]{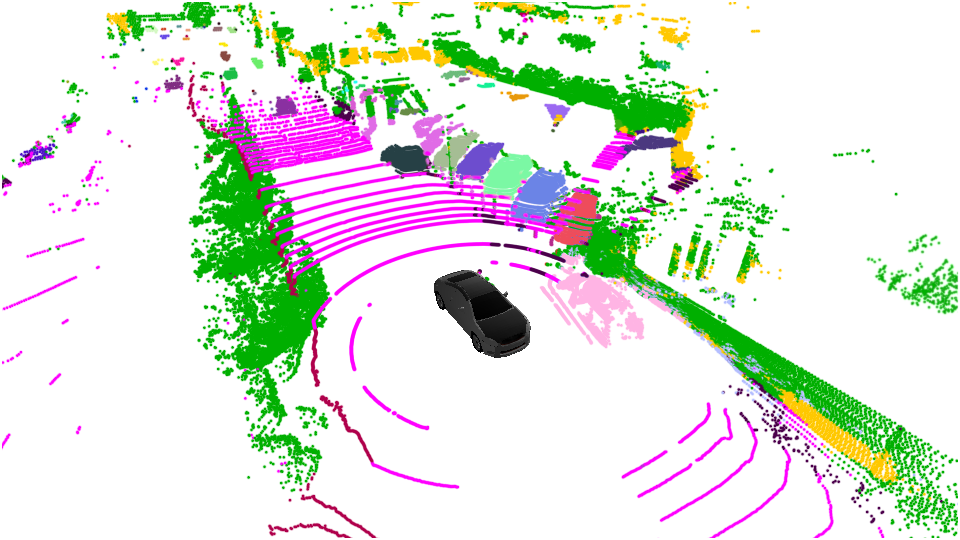}}
&
{\includegraphics[width=\linewidth, frame]{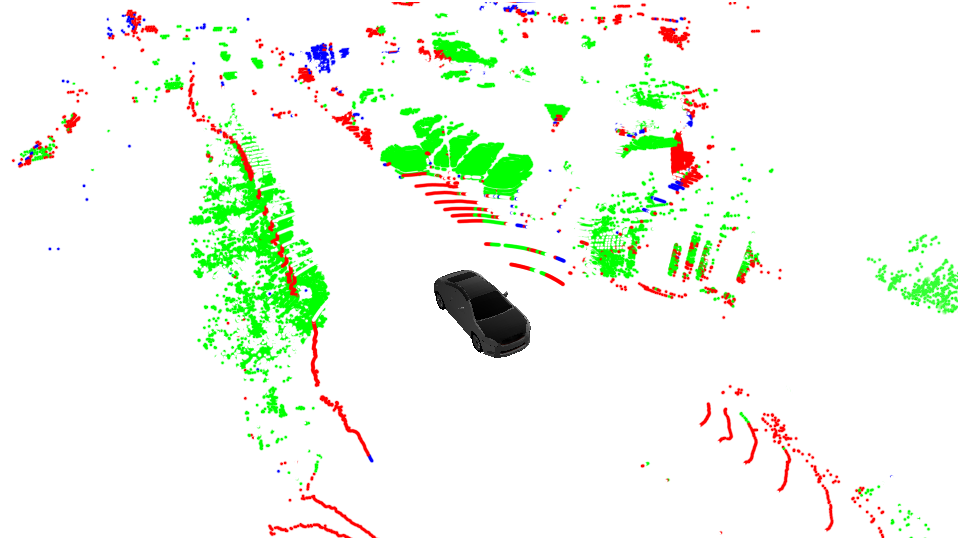}}
\\
\end{tabular}
}
\caption{
Qualitative results of \net\ in comparison with the best performing baseline from \tabref{tab:panoptic-da} on three dataset pairs: SemanticKITTI to nuScenes (K $ \rightarrow $ N), SemanticKITTI to PandaSet (K $ \rightarrow $ P), and nuScenes to PandaSet (N $ \rightarrow $ P). All UDA methods are employed on the EfficientLPS network. The third column shows the Improvement/Error map which depicts points misclassified by the baseline and correctly predicted by the \net\ model in green, points misclassified by \net\ and correctly by the baseline in blue, and points misclassified by both models in red.}
\label{fig:qual-analysis}
\vspace{-5.5mm}
\end{figure*}

In this section, we study the impact of the various components of our domain adaptation strategy in the form of an ablation study. \tabref{tab:network-ablation} presents the results of this study on the SemanticKITTI $\rightarrow$ nuScenes dataset pair. 
Model M1 consisting of the vanilla EfficientLPS model acts as the base model upon which we progressively add the subsequent data-based and model-based domain adaptation strategies. This base model achieves a PQ score of $10.27\%$. Upon employing the pose correction module in model M2, we observe an improvement of \SI{1.70}{pp} which can be attributed to the reduction in distribution disparity of the range, $x, y, \text{and}\ z$ channels between the source and target domains. We account for the disparity in the intensity models between the domains by introducing IMN in model M3 which results in a further improvement of \SI{1.50}{pp} in the PQ metric. In model M4, we introduce PDC-Lite to minimize the domain gap introduced by the batch normalization layers. PDC-Lite re-calibrates the mean and variance of the first batch normalization layer using the target dataset which allows it to achieve a PQ value of $14.25\%$. We introduce our core model-based adaptation strategy, MS-FSOT, in model M5 where we minimize the domain gap between the feature spaces of the source and target domains using an unbalanced optimal transport formulation. This addition results in a significant \SI{8.94}{pp} improvement in the PQ score, thus highlighting the importance of feature space alignment during domain adaptation. Finally, we employ our novel IAS in model M6 to ensure equal sampling importance for all classes and instances in the optimal transport cost matrix. This strategy results in a further \SI{1.81}{pp} improvement in the PQ score. 
We denote model M6 as our proposed \net\ domain adaption sampling strategy, as it achieves the best performance.
\vspace{-2mm}

\subsection{Qualitative Evaluation}
\label{subsec:qualitative-evaluation}

\figref{fig:qual-analysis} presents the qualitative results from EfficientLPS + \net\ on all three dataset pairs as compared to EfficientLPS without domain adaptation and the best baseline from \tabref{tab:panoptic-da}.
We also present an Improvement/Error Map for each example to highlight the points where the predictions of \net\ and the best baseline disagree to aid visual understanding. We observe from a large number of green areas in the Improvement/Error maps that \net\ effectively reduces the domain gap between the datasets resulting in significantly better panoptic segmentation outputs.
Further, we note from \figref{fig:qual-analysis} that EfficientLPS without domain adaptation struggles to predict the under-represented classes and often makes incorrect class predictions. This is evident in the first column of \figref{fig:qual-analysis}(a) and \figref{fig:qual-analysis}(b) where the model predicts cars and sidewalk\colsq{sidewalk} as building\colsq{building}, and road\colsq{road} and building as other-vehicle\colsq{other-vehicle} respectively.
The best baseline preserves the sharp boundaries between the semantic and instance predictions but often incorrectly predicts the \textit{stuff} classes as well as misclassifies \textit{thing} instances as \textit{stuff} classes. In contrast, our \net\ domain adaptation strategy generates significantly better predictions as compared to the best baseline, especially with respect to the detection of instances. This is most evident from the second and third columns of \figref{fig:qual-analysis}(c)-(f) where the baseline often incorrectly predicts vegetation\colsq{vegetation} as building, and cars as road or vegetation. Our model accurately handles such scenarios, and also correctly detects and differentiates between different instances in the scene, depicted by the differently colored vehicles in these images, even in extremely challenging scenarios.\looseness=-1

\section{Conclusion}
\label{sec:conclusion}

In this paper, we present the first end-to-end trainable unsupervised domain adaptation approach for LiDAR panoptic segmentation. Our \net\ consists of two separate domain adaptation strategies, namely, data-based domain adaptation and model-based domain adaptation. Data-based domain adaptation reduces the domain gap in the data using pose correction to account for the different LiDAR mounting positions and orientations, and virtual scan generation to simulate target-like data using the source data. Moreover, we perform intensity mapping using a learnable Intensity Mapping Network to map the intensity model of the target data to the source data. Model-based domain adaption addresses the domain gap between the panoptic segmentation models of the source and target domains using our novel MS-FSOT formulation augmented with IAS. MS-FSOT avoids various assumptions about the data distributions and effectively aligns the distributions even in the presence of extreme outliers. IAS augments MS-FSOT by intelligently sub-sampling the intermediate features while preserving both the local and global structures of the underlying distributions using panoptic label information. Using extensive evaluations on three different dataset pairs, we demonstrate that our \net\ strategy consistently outperforms existing domain adaptation techniques on both panoptic and semantic segmentation tasks. Lastly, we present a detailed ablation study along with a comprehensive qualitative analysis that highlights the improvement brought about by the different constituent elements of our domain adaption strategy.

\section*{Acknowledgments}

This work was funded by the Eva Mayr-Stihl Stiftung and the Federal Ministry of Education and Research~(BMBF) of Germany under ISA 4.0.

\footnotesize
\bibliographystyle{IEEEtran}
\bibliography{references.bib}

\end{document}